\documentclass{article}

\usepackage{PRIMEarxiv}

\usepackage{amssymb}
\usepackage{amsmath}
\usepackage{float}
\usepackage{xcolor}
\usepackage{xspace}
\usepackage{booktabs}
\usepackage{makecell}
\usepackage{multirow}
\usepackage{verbatim}
\usepackage{subcaption}
\usepackage[colorlinks=true,linkcolor=blue]{hyperref}
\usepackage[linesnumbered,ruled,vlined]{algorithm2e}
\usepackage{enumerate}
\usepackage[shortlabels]{enumitem}

\usepackage{xurl}
\usepackage{helvet}
\usepackage{etoolbox}
\usepackage{graphicx}
\usepackage{titlesec}
\usepackage{caption}
\usepackage{scrextend}
\usepackage{amsfonts}
\usepackage{bm}
\usepackage{amsmath}
\usepackage{soul}

\usepackage{tikz}
\relax
\usepackage[edges]{forest}
\usetikzlibrary{shadows.blur}

\usepackage{array}
\usepackage{tabularx}
\usepackage{ragged2e}
\usepackage{threeparttable}
\usepackage{lscape} 

\usepackage{float}
\floatstyle{plaintop}
\restylefloat{table}
\title{Decision-Focused Learning Enhanced by Automated Feature Engineering for Energy Storage Optimisation}

\author{
\begin{tabular}{c}
  Nasser Alkhulaifi$^{1}$, Ismail Gokay Dogan$^{1}$, Timothy R. Cargan$^{1}$, \\
  Alexander L. Bowler$^{2}$, Direnc Pekaslan$^{3}$, Nicholas J. Watson$^{2}$, \\
  Isaac Triguero$^{4,5,1}$ \\
\end{tabular} \\
$^1$ School of Computer Science, University of Nottingham, Nottingham, NG8 1BB, UK \\
$^2$ School of Food Science and Nutrition, University of Leeds, Leeds, LS2 9JT, UK \\
$^3$ LUCID Lab, School of Computer Science, University of Nottingham, Nottingham, NG8 1BB, UK \\
$^4$ Department of Computer Science and Artificial Intelligence, University of Granada, Granada, Spain \\
$^5$ DaSCI Institute, University of Granada, Granada, Spain \\
\texttt{Corresponding authors: nasser.alkhulaifi@nottingham.ac.uk, triguero@decsai.ugr.es}
}

\begin{document}
\maketitle

\begin{abstract}
Decision-making under uncertainty in energy management is complicated by unknown parameters hindering optimal strategies, particularly in Battery Energy Storage System (BESS) operations. Predict-Then-Optimise (PTO) approaches treat forecasting and optimisation as separate processes, allowing prediction errors to cascade into suboptimal decisions as models minimise forecasting errors rather than optimising downstream tasks. The emerging Decision-Focused Learning (DFL) methods overcome this limitation by integrating prediction and optimisation; however, they are relatively new and have been tested primarily on synthetic datasets or small-scale problems, with limited evidence of their practical viability. Real-world BESS applications present additional challenges, including greater variability and data scarcity due to collection constraints and operational limitations. Because of these challenges, this work leverages Automated Feature Engineering (AFE) to extract richer representations and improve the nascent approach of DFL. We propose an AFE–DFL framework suitable for small datasets that forecasts electricity prices and demand while optimising BESS operations to minimise costs. We validate its effectiveness on a novel real-world UK property dataset. The evaluation compares DFL methods against PTO, with and without AFE. The results show that, on average, DFL yields lower operating costs than PTO and adding AFE further improves the performance of DFL methods by 22.9–56.5\% compared to the same models without AFE. These findings provide empirical evidence for DFL's practical viability in real-world settings, indicating that domain-specific AFE enhances DFL and reduces reliance on domain expertise for BESS optimisation, yielding economic benefits with broader implications for energy management systems facing similar challenges.
\end{abstract}

\keywords{Decision-Focused Learning \and Predict-And-Optimise \and Predict-Then-Optimise \and Automated Feature Engineering \and Energy Storage Optimisations}


\section{Introduction and background}
\label{ch:Intro}

Decision-making under uncertainty is common in real-world applications where unknown parameters significantly complicate the process \cite{Ibrahim2020, Reza2023}. For example, in residential energy systems with Battery Energy Storage System (BESS), operators must make critical decisions about when to charge or discharge batteries to exploit time-varying tariffs (i.e., minimising electricity costs), and how much energy to store or release, while respecting physical and operational constraints of BESS \cite{Yang2022, Yu2023}. These decisions are made under uncertainty in both future electricity prices and household demand. In the literature, addressing these challenges typically involves a two-stage process: Machine Learning (ML) models forecast unknown variables, and then use these predictions as input parameters for Constrained Optimisation (CO) to determine optimal decisions within set boundaries \cite{Bergmeir2025}. This traditional sequential approach, also known as Predict-Then-Optimise (PTO), handles prediction and optimisation in isolation \cite{Vanderschueren2022}. The limitations of this approach manifest in two distinct ways: a) cascading errors arise from the sequential structure of PTO, where inaccuracies in the prediction stage may propagate and amplify through the optimisation stage, leading to suboptimal decisions \cite{Wilder2019, mandi2022_Through_the_lens}; and b) PTO suffers from objective misalignment in the utility of information extraction, as it focuses on deriving features and patterns from data solely to minimise prediction errors, without prioritising the information most relevant or useful for the downstream decision task \cite{donti2017task, Boettiger2022}.

To overcome this challenge, an emerging approach known as Decision-Focused Learning (DFL) integrates predictive modelling and optimisation directly into the learning process \cite{Wilder2019}. In this approach, forecasts are selected or assessed based on their impact on the actual downstream cost of the optimisation problem, rather than on standard error metrics (i.e., error-based loss function) such as Mean Squared Error (MSE). In order to achieve this, a task-aware loss function, such as regret, can be used \cite{Mandi2020}. By incorporating regret-based loss functions, which measure the difference between realised outcomes under uncertainty and optimal outcomes under perfect foresight, DFL methods have the potential to enhance decision quality by aligning training with the end-use task and prioritising it over minimising forecasting errors \cite{AnisLahoud2025}.

In gradient-based DFL literature, Smart ``Predict, then Optimise'' (SPO$^{+}$) \cite{Elmachtoub2022} and Differentiable Black-Box (DBB) \cite{pogan2020} are two seminal methods. In brief, the SPO$^{+}$ method employs a convex surrogate loss function, derived via duality theory, that upper bounds the SPO loss, which measures the decision error (suboptimality gap) induced by predicted cost vectors in a linear, convex, or integer optimisation problem, enabling efficient gradient-based training tailored to optimise decision quality. DBB method implements an efficient backward pass for blackbox combinatorial solvers with linear objective functions by constructing a continuous interpolation function, whose gradient is computed using a single solver call on perturbed inputs, enabling the integration of combinatorial algorithms into neural network architectures. These methods have been applied to various classical DFL problems, such as the travelling salesman and shortest path problems \cite{pogan2020}. However, despite promising theoretical advances and growing interest in DFL research, these methods have predominantly been evaluated on synthetic benchmark problems, with a lack of real-world applications \cite{kotary2021end, Mandi2024, geng2024benchmarking}. Therefore, DFL methods need to be explored beyond small-scale synthetic problems (referred to as \textit{toy-level} problems in \cite{Mandi2020, Zhou2024}) to demonstrate their practical viability using real-world data and constraints.

Whilst there are a few recent attempts to use DFL in real-world settings with different dataset sizes (e.g., multi-year \cite{Bergmeir2025}, one year \cite{Paredes2025}, two years \cite{Wang2025_AI_Optimized}, and six years \cite{Sang2022_6_years}), these datasets are significantly larger than those found in many practical applications where data scarcity is common due to collection limitations, privacy concerns, or resource constraints \cite{grinsztajn2022tree, hollmann2022tabpfn, Alkhulaifi2024_pipeline}. This highlights a critical gap in evaluating DFL's performance under constrained, small-scale real-world conditions, such as the 55-day dataset used in this study. In such scenarios, Feature Engineering (FE) can compensate for limited data by extracting informative features, thus maximising the utility of the available data as well as enhancing computational efficiency by eliminating noisy or irrelevant data \cite{Wang_2022}. However, manual FE for energy forecasting remains a time-consuming process that is prone to human error while relying heavily on domain expertise and iterative experimentation \cite{Wang_2022, Wu2022}. As a result, tasks such as FE and the integration of domain knowledge are largely left to human practitioners, which in turn has led to growing interest in Automated FE (AFE) methods \cite{hollmann2024large}. While AFE traditionally focuses on improving energy forecasting error metrics \cite{Alkhulaifi2025}, it remains unclear whether such improvements yield better operational outcomes when integrated with DFL approaches.

As the integration of renewables accelerates, prediction and optimisation approaches have attracted growing attention. For instance, as part of the IEEE-CIS Technical Challenge \cite{Bergmeir2025}, participants used different methods to forecast 15-minute power demand and solar production for six buildings and six solar arrays, while the optimisation task was to generate a schedule for a set of activities that minimised electricity costs across the buildings. Building on this, \cite{abolghasemi2022predict} reported strong positive Pearson correlations (0.81-0.9) between forecasting accuracy and optimisation cost across overforecast and underforecast scenarios (i.e., perturbed). Yet, they found that this correlation is asymmetric, meaning unequal effects between overforecasting and underforecasting, and that forecast‑accuracy metrics may be sub‑optimal for minimising complex optimisation costs.

Similarly, the need for aligning optimisation with forecasting is particularly important in domains such as BESSs \cite{Yang2022, Yu2023}, where such problems serve as a testing ground for evaluating the interplay between forecasting accuracy and downstream decision quality. In such energy scheduling problems, where future energy demand and costs are unknown, prediction errors can significantly affect downstream decisions \cite{Mandi2020Interior}, highlighting the limitations of traditional PTO methods and underscoring the potential of DFL to yield more practically valuable outcomes. In various contexts, literature on BESSs has studied PTO methods \cite{Hannan2021, Song2024}, while only recently have a few studies investigated DFL-based approaches for reserve-market participation \cite{Paredes2025} and for developing bidding strategies for microgrids in day-ahead electricity markets \cite{Alrasheedi2024}. 

This work therefore addresses gaps that challenge effective BESS optimisation, stemming from: A) DFL methods promise to align prediction with downstream objectives \cite{Wilder2019, Mandi2020}; however, they are relatively new and have been tested primarily on synthetic datasets or small-scale problems (i.e., simplified benchmarks) \cite{mandi2023towards, Zhou2024}, highlighting the need to assess their practical viability in real-world applications such as BESS problems; and B) real-world datasets often exhibit greater variability and data scarcity due to practical constraints \cite{grinsztajn2022tree, hollmann2022tabpfn} which can compromise DFL performance and necessitate enhanced feature representations to extract richer information from limited data without the need for domain expertise. The novelty and contributions of this work are summarised as follows\footnote{\scriptsize{To support transparency and reproducibility, the historical electricity price and weather data, along with the code used for the experiments in this work, are publicly available on GitHub. See \protect\url{https://github.com/Nasser-Alkhulaifi/DFL}}}:

\begin{itemize}[leftmargin=15pt, itemsep=-2pt]

    \item Proposing a decision-aware, end-to-end ML forecasting and optimisation framework for BESS problems and is suitable for small dataset sizes. Rather than treating forecasting and optimisation as separate tasks, the framework uses DFL to jointly forecast electricity demand and prices while optimising BESS operations using a regret-based objective.
    
    \item Improving the nascent approach of DFL by leveraging domain-specific AFE to extract richer representations without reliance on domain expertise, as presented in our previous work \cite{Alkhulaifi2025}, thereby streamlining the development of DFL pipelines for BESS problems.
    
    \item Using novel real-world data collected from a UK-based property to evaluate the proposed framework with a comprehensive comparative analysis of PTO versus DFL approaches, with and without AFE, thereby demonstrating the practical viability of DFL methods in real-world BESS applications.

\end{itemize}

The structure of the remaining sections of this paper is as follows: \autoref{ch:Method} outlines the methods used for the BESS problem in this work, \autoref{ch:Experimental_design} provides details of the experimental design, including the datasets used and evaluation criteria. Analysis and discussion of findings are presented in \autoref{ch:Results_and_Discussion}. To conclude, \autoref{ch:Conclusion} summarises key insights.
\section{Methods}
\label{ch:Method}

This section outlines the problem under investigation in \autoref{sec:problem_definition}, presents the mathematical model for the BESS optimisation problem in \autoref{Mathematical_model}, and describes the prediction methods in \autoref{Pred_model}.


\subsection{Problem Definition}
\label{sec:problem_definition}

\begin{landscape}
\begin{figure}
\centering
\includegraphics[width=1\linewidth]{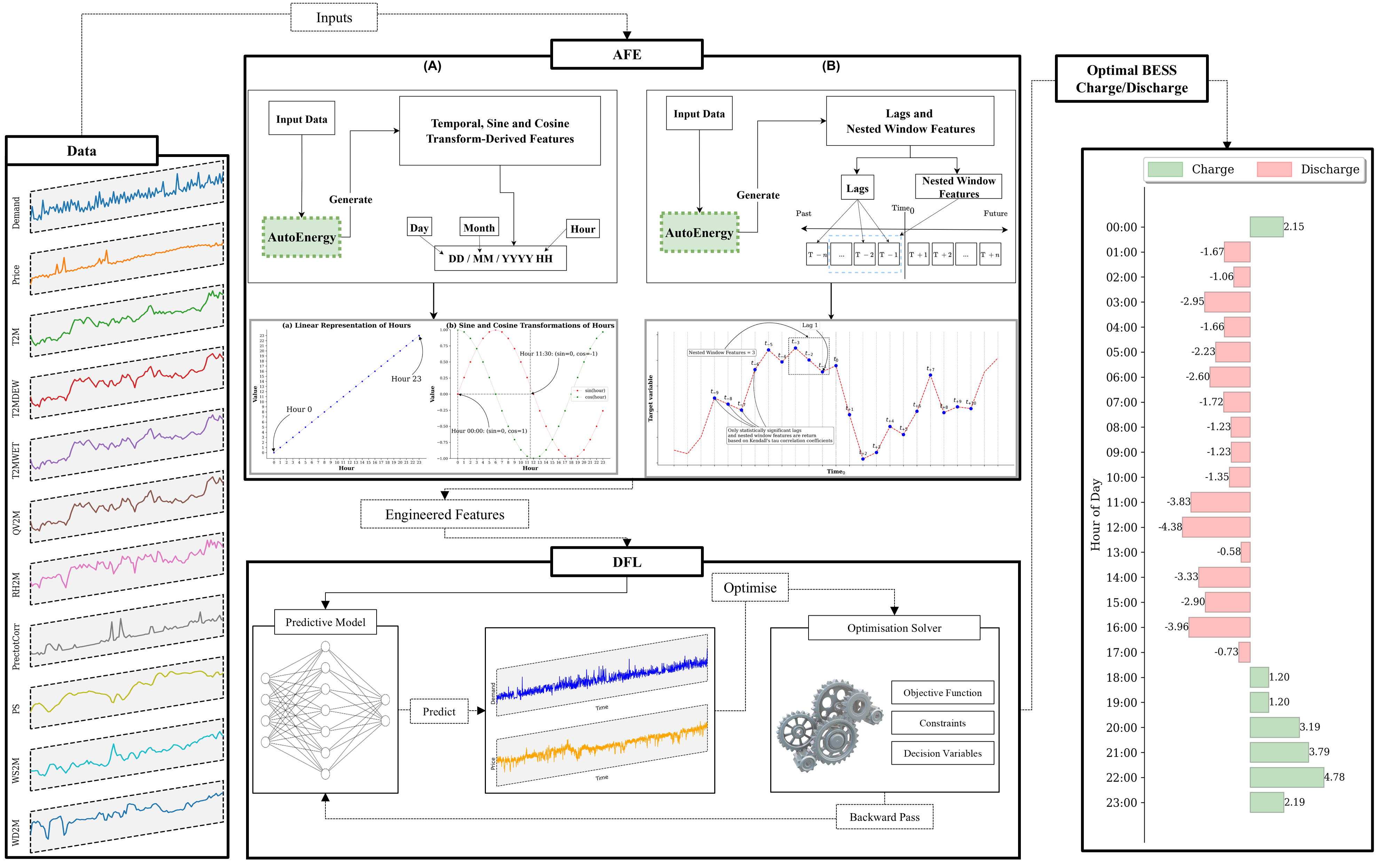}
\caption{The proposed framework jointly forecasts electricity prices and demand for the next day while optimising the downstream BESS task. It leverages AFE to enrich the dataset with domain-specific features using the \textit{AutoEnergy} algorithm \cite{Alkhulaifi2025}. Further details on this algorithm, including how the features are extracted and selected, are provided in \autoref{Auto_FE}. This approach aims to minimise domain knowledge requirements and enhance decision quality (i.e., improve the nascent approach of DFL) for BESS scheduling problems under data scarcity. It is worth noting that while the problem formulation presented in \autoref{ch:Method}, particularly the mathematical formulation in \autoref{Mathematical_model}, is tailored specifically for BESS applications, the overarching framework methodology is adaptable to other energy management contexts. The core principles of jointly forecasting uncertain parameters while optimising operational decisions remain applicable, though the specific constraints, decision variables, and objective functions would require modification for different energy systems.}
\label{Framework}
\end{figure}
\end{landscape}

The BESS problem entails forecasting unknown parameters (electricity prices and household demand), which serve as inputs to an optimisation model that computes the optimal charge/discharge schedule for the battery over the planning horizon while minimising electricity costs and satisfying energy demand and operational constraints. As illustrated in \autoref{Framework}, the proposed framework leverages AFE to enrich the dataset with domain-specific features \cite{Alkhulaifi2025} while reducing domain knowledge requirements, thereby enhancing decision quality and improving the nascent approach of DFL for BESS scheduling problems.

\subsubsection{Multiple Unknowns, Decision-Making, and Objectives}
In this study, the BESS problem involves predicting two key unknowns (electricity demand and price) over a 24-hour planning horizon, mimicking real-world residential energy management where household BESS uses day-ahead predictions to optimise charging and discharging schedules. These uncertainties can be framed as: \textbf{(A)} What will the electricity prices be over the next 24 hours? and \textbf{(B)} What levels of electricity demand are expected? Given these forecasts, the core decision problem is to determine the optimal battery operation strategy across the 24-hour horizon, which can be formulated through the following operational questions: \textbf{(C)} At what times (i.e., hour of the day) should the battery be charged or discharged? and \textbf{(D)} By how much should it be charged or discharged? Successfully addressing these four prediction and operational questions enables the achievement of the following objectives:
\begin{itemize}
    \item Exploit price differentials through optimal battery charging and discharging schedules to reduce overall electricity costs (i.e., cost minimisation).
    \item Ensure household energy requirements are consistently met through appropriate combinations of grid power and battery discharge (i.e., demand satisfaction).
    \item Generate a battery operation strategy that respects all battery physical constraints, including power limits, energy capacity, and state-of-charge boundaries (i.e., constraint compliance).
\end{itemize}


\subsection{Mathematical model of the BESS optimisation problem}
\label{Mathematical_model}

In this work, the optimisation problem of the BESS is formulated as a Mixed Integer Linear Programming (MILP) problem. MILP is widely adopted for BESS optimisation because it can efficiently handle both continuous variables (e.g., power flows and state of charge) and discrete decisions (e.g., on/off states or charge/discharge modes), enabling accurate modelling of operational constraints and system logic with computationally tractable and optimal solutions \cite{Yang2022, Yu2023}. The parameters are defined in \autoref{tab:parameters} and decision variables presented in \autoref{tab:Decision_Variables}, followed by the objective function \autoref{objective_func} subject to the listed constraints.

Let \( T = \{1, \ldots, T\} \) denote the set of all time intervals. For each interval \( t \in T \), \( p_t \) and \( d_t \) represent the unit price of electricity (i.e., the cost of 1 kWh) and the energy demand, respectively. The parameters \( \mathit{max\_charge} \) and \( \mathit{max\_discharge} \) denote the maximum amount of energy that can be charged to or discharged from the battery in a given interval. The parameter \( e_0 \) indicates the initial battery level (in kWh). The decision variable \( e_t \) denotes the battery level at the end of interval \( t \). The decision variables \( g_t \), \( b_t \), and \( c_t \) represent the amount of energy supplied from the grid to the property, from the battery to the property, and from the grid to the battery, respectively. The binary variable \( z_t \) takes the value 1 if energy is supplied from the battery to the property in interval \( t \), and 0 otherwise. 

\begin{table}[!t]
	\begin{center}
        \footnotesize
		\caption{Notations}
		{\renewcommand{\arraystretch}{1.3}
		\label{tab:parameters}
		\begin{tabular}{l l l} 
			\textbf{Set} & \textbf{Definition} & \textbf{ }  \\
			\hline	
			$T$ & Set of (time) intervals  &  \\
			$max\_charge$ & Maximum energy that can be added to the battery in an interval & \\
			$max\_discharge$ & Maximum energy that can be drained from the battery in an interval & \\ 
			$p_t$ & Price of 1 kWh energy at interval $t$ & $\forall t \in T $ \\
			$d_t$ & Energy demand at interval $t$  & $\forall t \in T $ \\
			$e_0$ & Initial battery level (kWh)  & \\
                $E_{\max}$ & Usable battery capacity (kWh) & \\
                $SoC_{\min}$ & Minimum allowable state-of-charge (fraction of $E_{\max}$) & \\
			\hline
		\end{tabular}}
	\end{center}
\end{table}

\begin{table}[!t]
	\begin{center}
        \footnotesize
		\caption{Decision Variables}
		{\renewcommand{\arraystretch}{1.3}
		\label{tab:Decision_Variables}
		\begin{tabular}{l l l} 
			\textbf{Dec. Var.} & \textbf{Definition} & \textbf{ }  \\
			\hline	
			$e_t$  & Amount of energy (kWh) in the battery at the end of interval $t$ & \\
			$g_t$  & Amount of energy (kWh) provided from grid to the property at interval $t$   & \\
			$b_t$  & Amount of energy (kWh) provided from battery to the property at interval $t$   & \\
			$c_t$  & Amount of energy (kWh) charged to battery at interval $t$   & \\
			$z_t$  & 1, if energy is being sent from battery to the property at interval $t$ & \\
			& 0, otherwise & \\

			\hline
		\end{tabular}}
	\end{center}
\end{table}


\begingroup
\allowdisplaybreaks
\setlength{\jot}{10pt}
\begin{align}
	& \text{Min} \quad \sum_{t \in T} p_t \cdot (g_t + c_t) && \label{objective_func}\\
	& \text{Subject to} && \notag \\
	& g_t + b_t = d_t && \forall t \in T \label{demand_satisfaction}\\
	& e_t = e_{t-1} + c_t - b_t && \forall t \in T \label{energy_balance}\\
	& b_t \leq e_{t-1} && \forall t \in T \label{available_energy_use}\\
        & b_t \leq M * z_t && \forall t \in T \label{if_charge_then_z}\\
	& c_t \leq M * (1 - z_t) && \forall t \in T \label{if_discharge_then_no_charge}\\
	& c_t \leq max\_charge && \forall t \in T \label{max_charge}\\
	& b_t \leq max\_discharge && \forall t \in T \label{max_discharge}\\
        & e_t \leq E_{\max} && \forall t \in T \label{capacity_limit}\\
        & e_t \geq SoC_{\min}\,E_{\max} && \forall t \in T \label{min_soc}\\
        & g_t, b_t, e_t, c_t \in \mathbb{R}^+ && \forall t \in T  \label{real_ranges} \\
	& z_t \in \{0,1\} && \forall t \in T \label{binary_ranges}
\end{align}
\endgroup

The objective function \eqref{objective_func} minimises the total electricity cost. Constraint \eqref{demand_satisfaction} ensures that the energy demand is satisfied for all intervals \( t \in T \). Constraint \eqref{energy_balance} enforces energy flow conservation (i.e., updating the battery’s energy level based on charging and discharging). Constraint \eqref{available_energy_use} ensures that the energy drawn from the battery never exceeds the available energy at the beginning of the interval. Constraints \eqref{if_charge_then_z} and \eqref{if_discharge_then_no_charge} jointly ensure that the battery cannot be charged and discharged simultaneously (i.e., using a large positive constant \( M \) to enforce the binary logic of \( z_t \)). Constraints \eqref{max_charge} and \eqref{max_discharge} impose upper bounds on charging and discharging rates. Constraint \eqref{capacity_limit} limits the state of charge so that it never exceeds the battery’s usable capacity \(E_{\max}\). Constraint \eqref{min_soc} enforces a floor of \(SoC_{\min}\,E_{\max}\) that prevents deep cycling, thereby reducing degradation and safeguarding long-term operational longevity. Constraints \eqref{real_ranges} and \eqref{binary_ranges} define the domains and the ranges of the decision variables.

\subsection{Prediction methods of the BESS forecasting problem}
\label{Pred_model}

Three Artificial Neural Network (ANN)-based methods (PTO, SPO$^{+}$, DBB) are used for the BESS forecasting problem in this work, based on the following methodological rationale. First, the conventional PTO method serves as the established baseline, representing the predominant approach in BESSs where prediction and optimisation are handled separately \cite{Vanderschueren2022}. Second, two DFL methods, namely SPO$^{+}$ \cite{Elmachtoub2022} and DBB \cite{pogan2020}, as introduced in \autoref{ch:Intro}, are selected as the two seminal and most widely adopted DFL methods in gradient-based DFL literature. This selection enables a comprehensive comparison between the conventional PTO approach and the two main DFL methods, thereby addressing the research gap regarding DFL's practical viability in real-world BESS applications.
 
The DFL-based methods are centred on training predictive models in an end-to-end approach using decision loss derived from the associated optimisation task, and therefore, an optimisation problem needs to be solved in each training iteration. In short, the SPO$^{+}$ method uses a convex surrogate loss to upper-bound decision regret in downstream optimisation tasks, while the DBB method enables end-to-end training by approximating solver outputs for gradient computation. Further details on loss computation for the SPO$^{+}$ and DBB methods can be found in \cite{Elmachtoub2022} and \cite{pogan2020}, respectively, and an explanation of the Python implementation is provided in \cite{Tang2024}. In contrast, the PTO method trains the predictive model with MSE loss, which measures the average squared difference between the model’s forecasts and the actual values, as shown in \autoref{MSE}. This approach, therefore, focuses solely on minimising forecasting errors during training, without considering the downstream optimisation task.
\begin{equation}
\operatorname{MSE} = \frac{1}{N}\sum_{i=1}^{N}\bigl(\hat{y}_{i} - y_{i}\bigr)^{2}
\label{MSE}
\end{equation}
where \(\hat{y}_{i}\) is the model's prediction for the \(i^{\text{th}}\) observation, \(y_{i}\) denotes the true value for that observation, and \(N\) is the total number of observations. 

\subsection{Automated Feature Engineering}
\label{Auto_FE}
This study leverages an AFE method specifically designed for energy forecasting problems, which we introduced in our previous work \cite{Alkhulaifi2025}. This method, named \textit{AutoEnergy}, aims to streamline ML pipeline development by automatically generating features from timestamps and historical energy consumption data, thereby minimising reliance on domain expertise for FE. It generates two primary categories of features: A) Temporal features: these features extract time-based patterns from timestamps such as hour of the day, day of the week, weekdays and weekends. To enhance the representation of cyclical patterns, these temporal features undergo sine and cosine transformations. Specifically, Fourier-based transformations are applied to capture periodic patterns that represent daily and weekly cycles inherent in energy data. B) Lag and nested window features: these features capture temporal dependencies and multi-scale statistical characteristics within the energy time series data. The method computes statistically significant lags and rolling statistics (e.g., mean and standard deviation) using Kendall's tau correlation coefficient. This approach ensures that only meaningful temporal relationships are incorporated into the feature set. Additional details of this method are explained in \cite{Alkhulaifi2025}.

\section{Experimental design}
\label{ch:Experimental_design}
This section outlines the experimental design used in this work, including the datasets used in \autoref{datasets_subsection}, BESS configuration in \autoref{BESS_configuration}, experimental procedure in \autoref{Experimental_procedure}, and lastly the evaluation criteria in \autoref{Evaluation_metrics}.


\subsection{Datasets}
\label{datasets_subsection}
\begin{figure}[!t]
\centering
\includegraphics[width=1\linewidth]{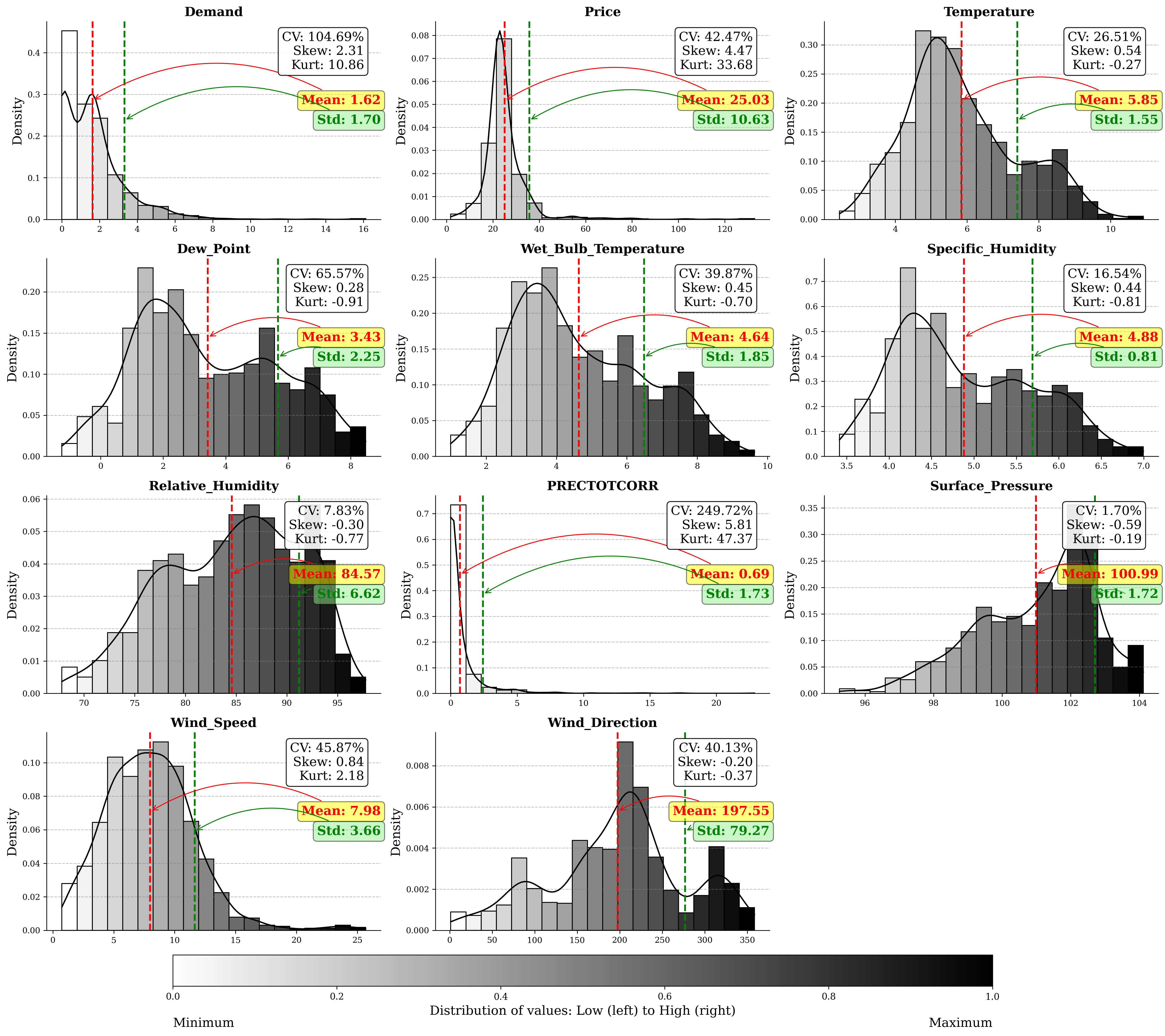}
\caption{The Dataset used in this study is depicted in histograms with colour-coded bars representing normalised bin positions. Red and green dashed lines indicate the mean and standard deviation, respectively. Annotated statistics include coefficient of variation (CV: relative variability), skewness (distribution asymmetry), and kurtosis (tailedness). The colour gradient in the histogram bars represents the distribution of values from low (left) to high (right).}
\label{fig:stats}
\end{figure}

\begin{figure}[!t]
\centering
\includegraphics[width=1\linewidth]{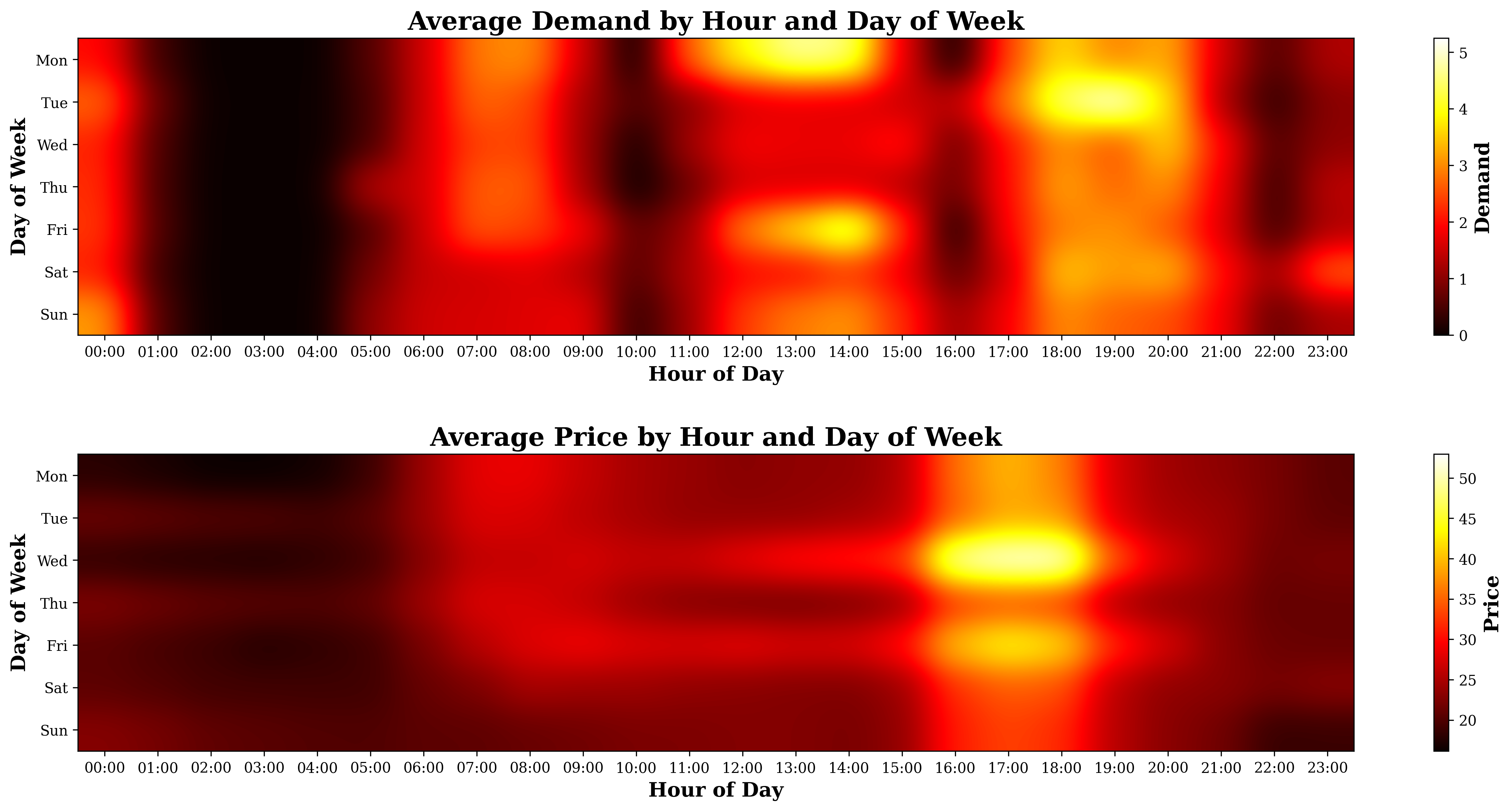}
\caption{Heatmaps of average hourly household electricity demand (top) and price (bottom) by day of week. Demand displays consistent peaks in the early morning (around breakfast), midday (lunch), and most prominently between 17:30 and 20:00 (dinner time), reflecting typical residential consumption patterns. Price peaks are concentrated between 16:00 and 19:30. Both demand and price are lowest during the early morning hours (01:00 to 05:00), reflecting reduced residential activity and system load during overnight periods.}
\label{fig:Demand_Price_Heatmaps}
\end{figure}

In this work, a dataset spanning 1 January 2025 to 24 February 2025 (55 days) is used. This scale is substantially smaller than in related works (e.g., multi-year datasets \cite{Bergmeir2025}, one year \cite{Paredes2025}, two years \cite{Wang2025_AI_Optimized}, and six years \cite{Sang2022_6_years}), thereby distinguishing the study’s contribution further by rigorously evaluating DFL, enhanced by AFE, under limited real-world data. This hourly historical real-world electricity demand and BESS data were provided by the Intelligent Plant platform \cite{IntelligentPlant}. Historical electricity prices were sourced via the Octopus Energy API \cite{OctopusEnergy}. Weather data was obtained from the NASA Langley Research Center's POWER Project \cite{NASAPOWER}, a repository of solar and meteorological datasets developed by NASA to support renewable energy and building energy efficiency research. \autoref{fig:stats} presents summary statistics of the collected data, while \autoref{fig:Demand_Price_Heatmaps} displays patterns of average hourly electricity demand and price by day of the week.


\subsection{BESS configuration}
\label{BESS_configuration}
\begin{figure}[!t]
\centering
\includegraphics[width=1\linewidth]{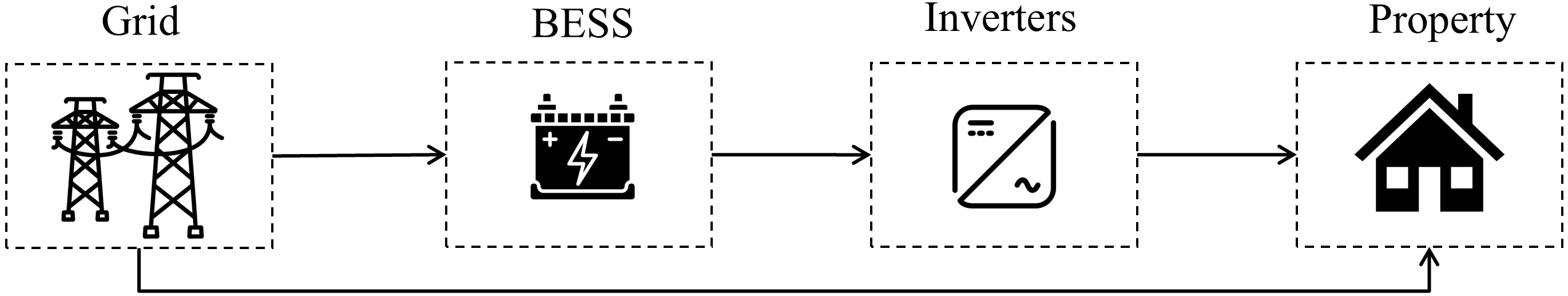}
\caption{Power flow diagram, where: A) power flows from the grid to charge the battery, which then supplies the property via inverters performing DC–AC conversion to meet energy demand; and B) power flows directly from the grid to the property when the battery capacity is insufficient.}
\label{fig:Power_flow}
\end{figure}

In this experiment, the maximum charging rate was set to 5 kWh per interval to comply with operational battery constraints, while the maximum discharging rate was limited to 4.5 kWh per interval to account for round-trip efficiency losses and system resistive losses inherent in BESSs. The minimum state of charge was maintained at 10\% to prevent battery degradation and ensure operational longevity. The capacity was set to 50 kWh, imposed by physical battery constraints. It is worth noting that the current BESS configuration operates without any solar panels. In this system, as shown in \autoref{fig:Power_flow}, electrical power follows two primary pathways: first, power flows from the grid to charge the battery, which then supplies the property through inverters that perform DC-AC conversion to meet energy demand; second, power flows directly from the grid to the property when battery capacity is insufficient. This dual-path configuration ensures a continuous power supply through direct grid connection while enabling BESS optimisation for energy arbitrage (i.e., storing electricity during low-cost periods and supplying the property during high-cost periods), thereby reducing energy cost and contributing to grid stability.


\subsection{Experimental procedure}
\label{Experimental_procedure}
In this work, the dataset is partitioned into three subsets: \(\sim\)50\% (27 days) allocated for initial training, \(\sim\)25\% (14 days) for validation to fine-tune hyperparameters, and the remaining \(\sim\)25\% (14 days) reserved for assessing model performance on previously unseen data (i.e., test dataset). ANN-based methods, as explained in \autoref{Pred_model}, were used to predict the price and demand for the next 24 hours (i.e., the following day) based on input features such as weather conditions (e.g., outdoor temperature) and the engineered features described in \autoref{Auto_FE}, while accounting for the operational and physical constraints of the BESS explained in \autoref{Mathematical_model}. This approach mimics real-world scenarios where BESS schedules charging and discharging operations daily to optimise cost savings by taking advantage of low electricity prices. It is important to note that, to better capture the interrelated dynamics between electricity price and demand, a multitask learning approach was used \cite{Wang2022}, in which the same input features are processed through shared ANN layers to predict both outputs simultaneously, thereby leveraging shared information across tasks.

\begin{figure}[!t]
\centering
\includegraphics[width=1\linewidth]{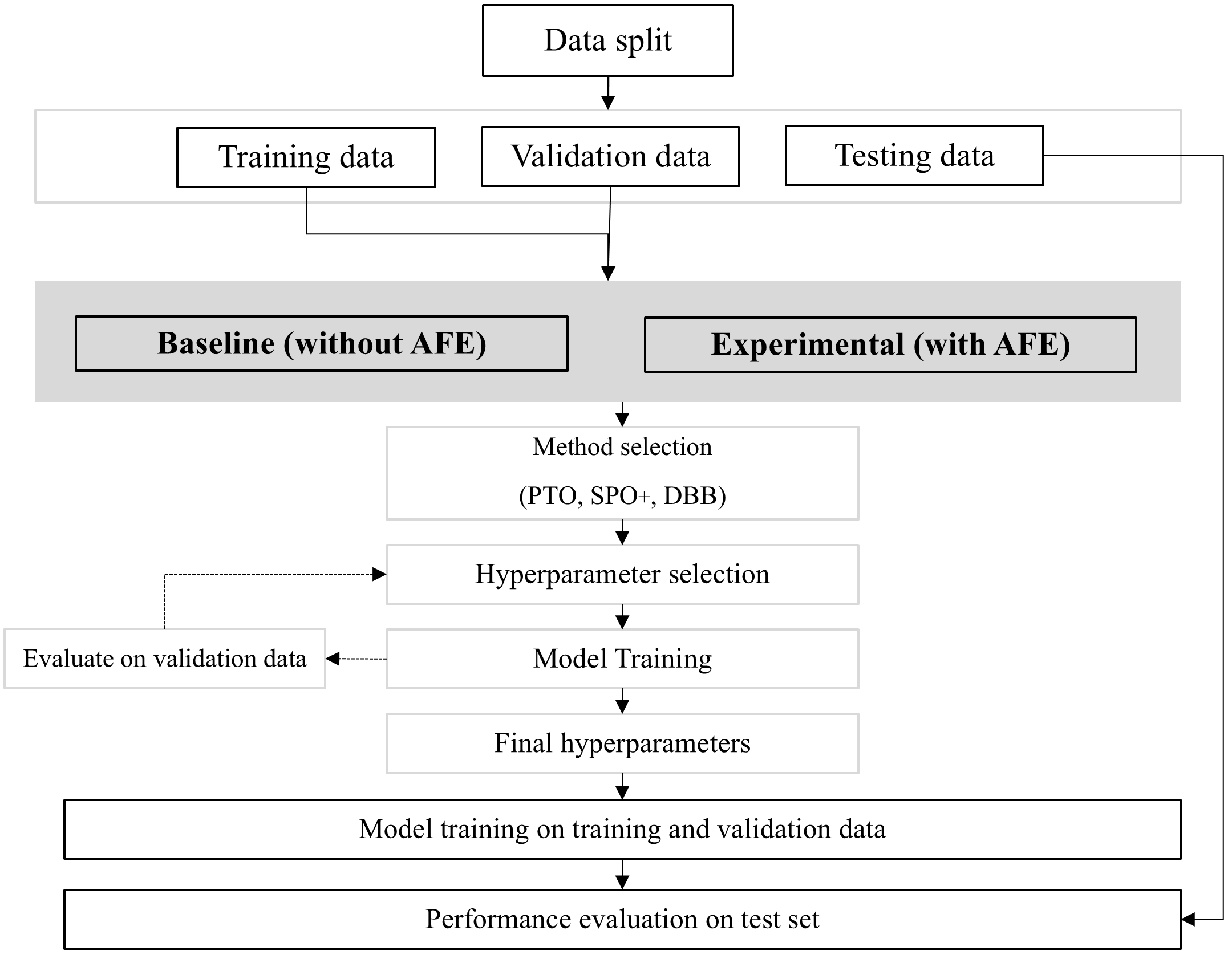}
\caption{Experimental design. See \autoref{Experimental_procedure} for a detailed explanation of the experimental procedure.}
\label{fig:experimental_design}
\end{figure}


Five-fold cross-validation and grid search optimisation were employed to tune the ANN architecture and learning hyperparameters of all three methods (PTO, DBB, and SPO$^{+}$). During the hyperparameter tuning phase, many model configurations are trained exclusively on the training set and evaluated on the validation set until optimal hyperparameters are identified, after which the training and validation sets are combined (totalling 75\%, 41 days of the original dataset) to train the final model using the previously optimised hyperparameters, thereby maximising the utilisation of available data for final model training. \autoref{fig:experimental_design} shows the experimental design and \autoref{tab:hyperparameters_space} shows the hyperparameter search space. Finally, the performance of the PTO, DBB, and SPO$^{+}$ methods was evaluated on the test set using the evaluation metrics described in \autoref{Evaluation_metrics}. To ensure the robustness and reliability of the experimental results, models were trained and evaluated across ten independent runs with different random seeds. This practice mitigates variance introduced by stochastic training procedures (e.g., such as random weight initialisation) and enables statistically meaningful comparisons by reporting mean performance metrics together with their standard deviations. The results of all ten independent runs are reported in the supplementary material. It is also important to note that the data were not shuffled, as the energy data exhibit temporal patterns, making it essential to preserve the chronological order of the timestamps.


\begin{table}[!t] 
\centering 
\begin{threeparttable}
\footnotesize

\setlength{\extrarowheight}{1pt} 
\begin{tabularx}{\textwidth}{
  @{}
  >{\centering\arraybackslash}m{3cm}
  >{\centering\arraybackslash}m{2.5cm}
  | >{\scriptsize\centering\arraybackslash}X
  | >{\scriptsize\centering\arraybackslash}X
  | >{\scriptsize\centering\arraybackslash}X
  | >{\scriptsize\centering\arraybackslash}X
  | >{\scriptsize\centering\arraybackslash}X
  | >{\scriptsize\centering\arraybackslash}X
  @{}
}

\toprule
\multirow{4}{*}{Hyperparameter} & \multirow{4}{*}{Search Space} & \multicolumn{6}{c}{Best Hyperparameter} \\ 
\cmidrule(lr){3-8} 
& & \multicolumn{2}{c!{\vrule width 0.8pt}}{PTO} & \multicolumn{2}{c!{\vrule width 0.8pt}}{SPO$^{+}$} & \multicolumn{2}{c}{DBB} \\ 
\cmidrule(lr){3-4} \cmidrule(lr){5-6} \cmidrule(lr){7-8} 
& & No AFE & AFE & No AFE & AFE & No AFE & AFE \\ 
\midrule 
Number of Layers & $\in \{1,\,2\}$ & 2 & 2 & 2 & 2 & 2 & 1 \\ 
Epochs & $\in \{10,\,20,\,30\}$ & 30 & 30 & 30 & 30 & 30 & 30 \\ 
Learning Rate & $\in \{10^{-3}, 10^{-5}\}$ & $10^{-3}$ & $10^{-3}$ & $10^{-5}$ & $10^{-5}$ & $10^{-3}$ & $10^{-3}$ \\ 
\bottomrule 
\end{tabularx} 
\caption{Hyperparameter search space and the corresponding best hyperparameters for the ANN models, determined through grid search on the validation dataset. To avoid over-parameterisation, the search space is constrained to the most critical parameters identified in related work \cite{Paredes2025}. Also, computational resource limitations further restrict the search space. The search includes architectures with one or two hidden layers; the best models use 256 and 128 neurons in the first and second hidden layers, respectively. The grid search explores 12 configurations with 5-fold cross-validation, yielding 60 candidate models per method and 180 ANNs in total (PTO, SPO\textsuperscript{+}, DBB); see \autoref{Pred_model} for details. For each training window, a 24-interval MILP (\autoref{Mathematical_model}) is solved using Gurobi \cite{gurobi} to optimise the battery schedule. During training, the solver is called once per sample (i.e., one day) per epoch; therefore, the optimisation layer dominates the total run time. This procedure is repeated twice (see \autoref{fig:experimental_design}), once with AFE and once without AFE, as explained in \autoref{Experimental_procedure}. Unlike \cite{Paredes2025}, which tunes core regressor hyperparameters on PTO and transfers them to DFL, we perform method-specific tuning: for PTO method, the models were trained with MSE loss on predictions; for DFL methods (SPO$^{+}$ and DBB), training integrates optimisation results via their respective decision-focused losses (see \cite{Mandi2020, Tang2024} for more details). The total running time, including hyperparameter tuning, is approximately 39.97 hours. In this experiment, models were trained using Python 3.11, Gurobi API \cite{gurobi}, and PyEPO API \cite{Tang2024}. Computational experiments were performed on an Ubuntu system featuring an \texttt{x86\_64} architecture with 16 physical CPU cores (32 logical cores), 64GB of RAM.}
\label{tab:hyperparameters_space} 
\end{threeparttable}
\end{table}

To evaluate the impact of AFE on decision quality, two conditions were established: A) a baseline scenario without AFE and B) an experimental scenario incorporating AFE. The baseline models serve to assess forecasting and optimisation capabilities with minimal inputs. This deliberately simplified configuration replicates model performance under conditions that simulate both a worst-case scenario (i.e., absence of domain-specific FE knowledge) and an initial testing phase where the model learns with limited data enhancement. Conversely, the experimental scenario applies the AFE method explained in \autoref{Auto_FE}, thereby enabling the assessment of the AFE impact on the downstream optimisation task (i.e., decision-making quality).


\subsection{Evaluation metric and statistical tests}
\label{Evaluation_metrics}

In this work, model performance is evaluated using the normalised regret metric \cite{Tang2024}. The notion of regret is used to measure the error in decision-making. It is characterised as the difference in the objective value between the true optimal solution and the optimal solution obtained by utilising the predicted coefficients. For minimisation problems, the normalised regret is defined as:

\begin{equation}
\label{eq:normalized_regret}
\frac{\sum_{i=1}^{n_{\text{test}}} \mathcal{L}_{\text{Regret}}(\hat{c}^i, c^i)}{\sum_{i=1}^{n_{\text{test}}} \lvert z^*(c^i) \rvert}
\end{equation}

where $\mathcal{L}_{\text{Regret}}(\hat{c}^i, c^i) = c^{i\top} w^*(\hat{c}^i) - z^*(c^i)$ denotes the regret for instance $i$. Here, $w^*(\hat{c}^i)$ represents the optimal solution obtained using the predicted cost vector $\hat{c}^i$, $c^{i\top} w^*(\hat{c}^i)$ is the objective value achieved by this solution when evaluated under the true cost vector $c^i$, and $z^*(c^i)$ is the true optimal objective value under the actual cost vector $c^i$. The regret thus measures the difference between the objective value of the solution derived from predicted costs (evaluated under true costs) and the true optimal objective value. The denominator normalises the aggregated regret by the sum of absolute values of true optimal objectives across the test set, providing a scale-invariant measure of decision quality degradation \cite{Tang2024}.

Non-parametric hypothesis tests were used to identify significant differences between the DFL and PTO methods, as well as between AFE versus without AFE and to support the experimental findings statistically \cite{Sheskin2003}. The Friedman Aligned-ranks test \cite{Garca2010} first evaluated overall differences among the methods with the significance threshold set at \(\alpha = 0.05\). Additionally, for pairwise comparisons, the Wilcoxon Signed-Rank test \cite{demvsar2006statistical, garcia2008extension} was applied at \(\alpha = 0.05\) to explore potential differences between method pairs that the preceding test did not flag as statistically significant, thereby ensuring a comprehensive statistical analysis.

\section{Results and Discussion}
\label{ch:Results_and_Discussion}

This section presents an analysis and discussion of the results.  
\autoref{Performance_overview} discusses the overall performance of the PTO and DFL methods for the investigated BESS problem. Subsequently, \autoref{FE_impact} provides a comparative analysis of the impact of AFE on the downstream optimisation task (i.e. decision quality) supported by feature importance analysis. Finally, \autoref{Statistical_Significance} evaluates the statistical significance of the observed performance differences between the methods, to verify that the reported results are unlikely to be due to chance. Supplementary materials present the detailed results of each of the ten runs across the test days, while \autoref{Appendix_A} illustrates the daily BESS optimisation schedules for the best run of each day using the best-performing method, showing how predicted prices and demand, subject to operational constraints, shaped the BESS scheduling decisions.

\subsection{Performance overview of PTO compared to DFL}
\label{Performance_overview}

\begin{figure}[!t]
\centering
\includegraphics[width=1\linewidth]{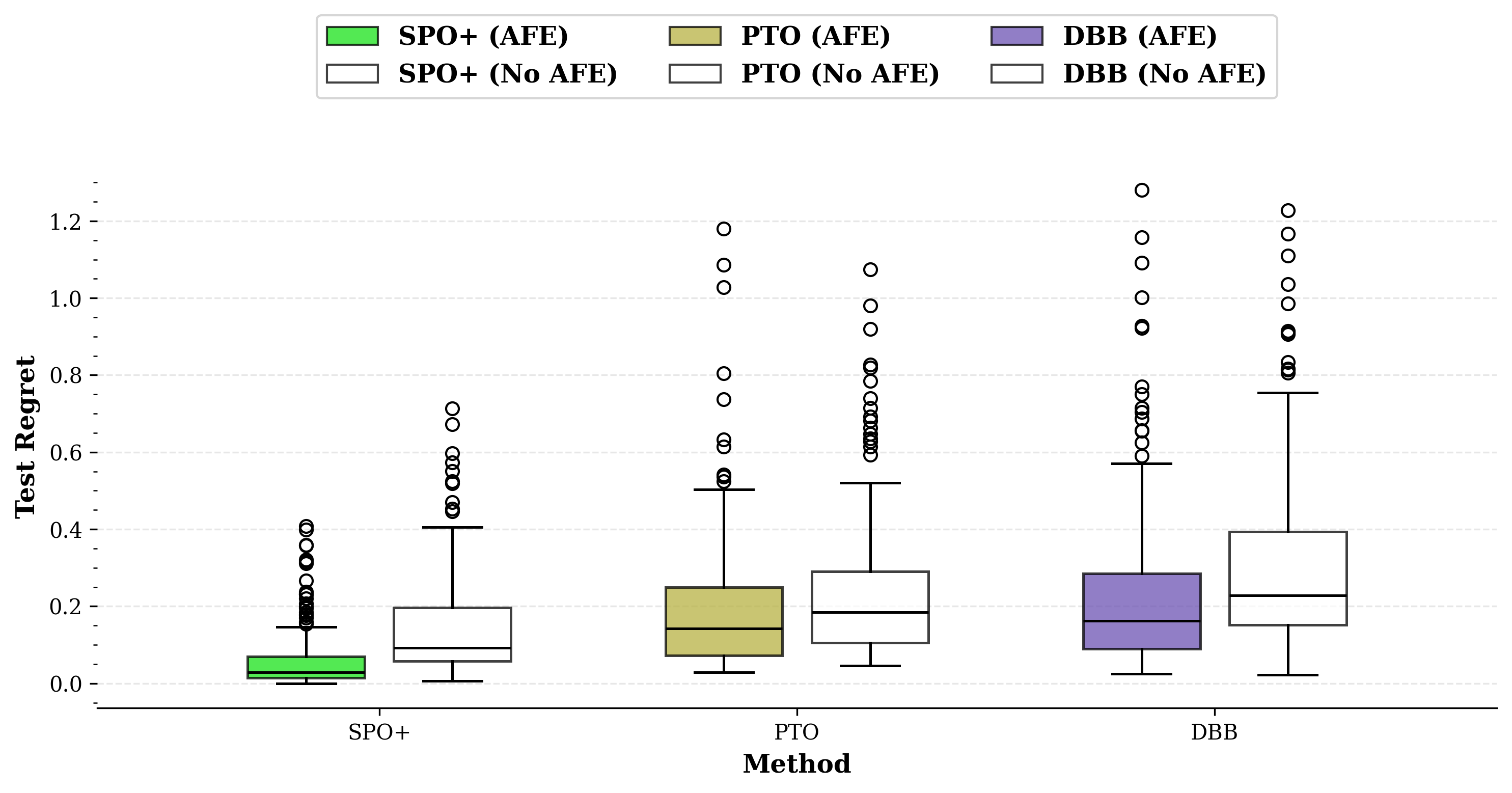}
\caption{Box plots showing the distribution of test regret across all ten experimental runs and test days, comparing PTO with DFL methods (SPO$^{+}$ and DBB), both with and without AFE. Each box displays the interquartile range (25th–75th percentiles) with the median marked by the central line, whiskers extend to the data range, and outliers are shown as individual points. The distributions reflect performance variability across multiple independent experimental runs and test days, providing a comprehensive view of method robustness and central tendency.}

\label{Box_plots}
\end{figure}

The performance comparison results between the PTO and DFL methods across the test set are presented in \autoref{Box_plots} and \autoref{tab:test_regret}.
Across the fourteen‑day test horizon, the SPO$^{+}$ method outperformed both the traditional PTO  and DBB methods. On average, the PTO method with AFE incurred a regret of 0.2046, while the SPO$^{+}$ method achieved a remarkably low regret of 0.0672. This improvement corresponds to a substantial reduction of approximately 67.16\%, underscoring the significant benefits of integrating the optimisation layer during training with the SPO$^{+}$. In other words, when the model is trained to predict electricity price and property demand while minimising downstream cost using the SPO$^{+}$ loss, rather than a generic pointwise error metric (i.e., error-minimising forecasts), it yields cheaper battery schedules. Interestingly, DBB, which is considered a DFL approach, achieved the highest average regret. This observation is consistent with recent related work where DBB underperforms SPO$^{+}$ \cite{zharmagambetov2023landscape, Wang2025_AI_Optimized}, despite differences in datasets and problem formulations. Although DBB can differentiate through black-box combinatorial solvers, the findings of this work indicate that this method is less suitable for BESS optimisation tasks under data scarcity, as evidenced by the higher regrets compared with other methods, potentially due to its original design for problems with linear objectives \cite{Mandi2024}.

An examination of day‑to‑day performance reveals that SPO$^{+}$ with AFE delivers notably consistent superior performance, demonstrating reliability on a per-instance basis. This consistency is reflected in lower variance (0.0780 vs 0.1732 for PTO and 0.2128 for DBB), as alternative approaches exhibited greater sensitivity to volatile market conditions such as those observed on February 22nd and 24th. The stability demonstrated by SPO$^{+}$ suggests that training with regret-based loss may produce models that are more robust to diverse price-demand patterns compared to both traditional PTO and DBB methods. Such findings, while data-dependent, indicate that SPO$^{+}$ could offer practical advantages for BESS strategies by potentially reducing exposure to costly high-regret scheduling decisions, though broader validation across different operational contexts would strengthen these conclusions.


\begin{table}[!t]
\centering
\footnotesize
\setlength{\extrarowheight}{1pt}
\begin{tabularx}{\textwidth}{ @{} >{\centering\arraybackslash}m{2.8cm} >{\centering\arraybackslash}X >{\centering\arraybackslash}X !{\vrule width 0.8pt} >{\centering\arraybackslash}X >{\centering\arraybackslash}X !{\vrule width 0.8pt} >{\centering\arraybackslash}X >{\centering\arraybackslash}X @{} }
\toprule
\multirow{4}{*}{Date} & \multicolumn{6}{c}{Test Set Regrets} \\
\cmidrule(lr){2-7}
 & \multicolumn{2}{c!{\vrule width 0.8pt}}{PTO} & \multicolumn{2}{c!{\vrule width 0.8pt}}{SPO$^{+}$} & \multicolumn{2}{c}{DBB} \\
\cmidrule(lr){2-3} \cmidrule(lr){4-5} \cmidrule(lr){6-7}
 & AFE & No AFE & AFE & No AFE & AFE & No AFE \\
\midrule
11 Feb & 0.0620 & 0.0780 & 0.0079 & 0.0611 & 0.0832 & 0.1247 \\
12 Feb & 0.1234 & 0.1669 & 0.0369 & 0.0845 & 0.1440 & 0.1858 \\
13 Feb & 0.1573 & 0.1968 & 0.0205 & 0.1222 & 0.1685 & 0.2243 \\
14 Feb & 0.1856 & 0.2363 & 0.0469 & 0.1736 & 0.2101 & 0.2678 \\
15 Feb & 0.0800 & 0.1030 & 0.0173 & 0.0571 & 0.1133 & 0.1335 \\
16 Feb & 0.0599 & 0.0681 & 0.0169 & 0.0496 & 0.0708 & 0.1042 \\
17 Feb & 0.0985 & 0.1384 & 0.0406 & 0.0880 & 0.1377 & 0.1818 \\
18 Feb & 0.1495 & 0.1939 & 0.0586 & 0.1142 & 0.1525 & 0.2711 \\
19 Feb & 0.1483 & 0.1784 & 0.0334 & 0.0891 & 0.1420 & 0.2156 \\
20 Feb & 0.1144 & 0.1530 & 0.0344 & 0.0909 & 0.1366 & 0.1862 \\
21 Feb & 0.2495 & 0.3261 & 0.0322 & 0.1395 & 0.3041 & 0.4344 \\
22 Feb & 0.4604 & 0.5771 & 0.1334 & 0.3199 & 0.5789 & 0.7216 \\
23 Feb & 0.2994 & 0.3363 & 0.2791 & 0.3367 & 0.3896 & 0.4549 \\
24 Feb & 0.6756 & 0.8021 & 0.1832 & 0.4343 & 0.8098 & 0.9551 \\
\midrule
Mean & 0.2046 & 0.2539 & \textbf{0.0672} & 0.1543 & 0.2458 & 0.3187 \\
Std & 0.1732 & 0.2055 & \textbf{0.0780} & 0.1206 & 0.2128 & 0.2474 \\
\bottomrule
\end{tabularx}
\caption{Daily test regrets (lower is better) averaged across ten independent runs to improve the robustness and reliability of the results (see supplementary materials for detailed results of each of the ten runs across the test set days). The lowest mean and standard deviation values are bolded to indicate the best performance. The comparison includes PTO and DFL methods (SPO$^{+}$, DBB), each evaluated with and without AFE (see \autoref{Experimental_procedure} for more details on the experimental procedure). Regret values represent the additional electricity cost incurred due to imperfect predictions when making battery scheduling decisions, where a regret of zero indicates optimal decision-making. Over the 14 days in the test set, SPO$^{+}$ consistently achieved lower regrets than the traditional PTO and DBB approaches, with AFE consistently improving performance across all methods. Appendix \ref{Appendix_A} provides detailed daily BESS optimisation schedules for the best run of each day using the best-performing method.}
\label{tab:test_regret}
\end{table}



\subsection{Impact of AFE on PTO and DFL Methods with Feature Importance Analysis}
\label{FE_impact}

The addition of AFE significantly enhanced the performance for all methods, evidenced by the results presented in \autoref{Box_plots} and \autoref{tab:test_regret}, with particularly pronounced benefits for the DFL approaches. Without AFE, SPO$^{+}$ outperformed the other methods with a mean regret of 0.1543, compared to the PTO approach (0.2539) and DBB (0.3187). This initial superiority suggests that the SPO$^{+}$ formulation is inherently effective at learning the mapping between inputs and optimal decisions for the investigated problem, even with less feature representations. However, the introduction of AFE amplified these advantages, with SPO$^{+}$ achieving 56.48\% improvement in mean regret, while the PTO improved by 19.42\% and DBB by 22.87\%. The overall reductions in regrets indicate that the AFE, while minimising the need for expert-driven FE, successfully generates useful features not only for maximising forecasting accuracy (i.e., the PTO approach) but also for optimising downstream decision-making tasks within DFL frameworks (i.e., SPO$^{+}$ and DBB). Moreover, the notable responsiveness of SPO$^{+}$ to enhanced feature representations may suggest the synergistic effect between this method and AFE in BESS optimisation problems. It implies that AFE provides SPO$^{+}$ with the contextual information necessary to better approximate the discrete optimisation problem via its convex surrogate, leading to more cost-effective battery scheduling decisions. The enriched feature representations could enable the convex surrogate loss to better capture the relationship between prediction errors and their downstream decision costs.


\begin{figure}[!t]
\centering
\includegraphics[width=1\linewidth]{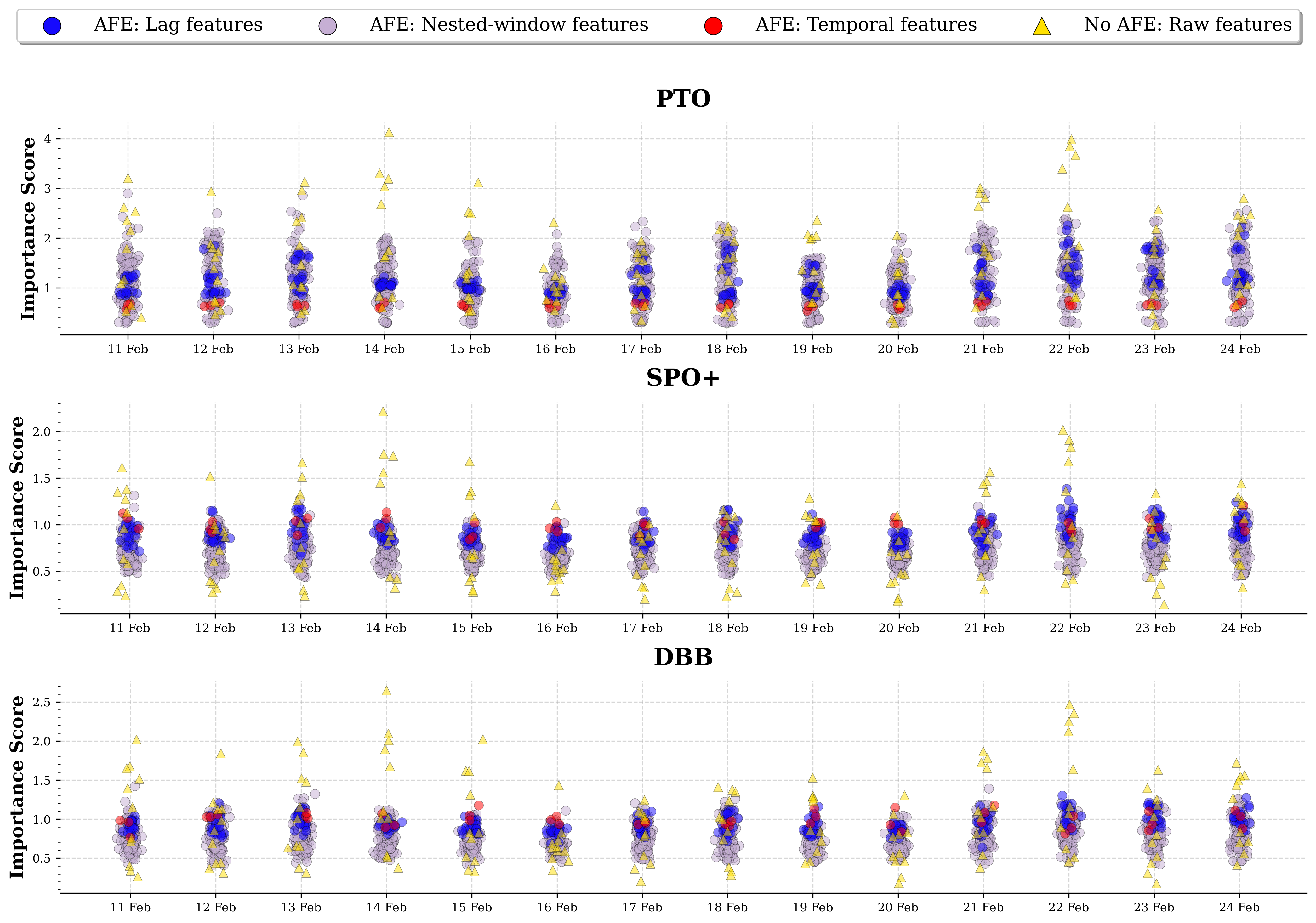}
\caption{Feature importance scores across all experimental runs for each method across test days. Each subplot corresponds to one method and shows the average SHAP-derived importance of individual features. Feature categories are distinguished by colour and marker shape. ‘No AFE’ indicates raw features, such as weather variables (e.g., outdoor temperature), while ‘AFE’ refers to automatically engineered features (e.g., lagged values, rolling-window statistics, temporal/calendar-based indicators). See \autoref{Auto_FE} and our previous work \cite{Alkhulaifi2024, Alkhulaifi2025} for further details on what these features represent and how they are generated and selected. Each marker represents a single feature from the corresponding category, allowing the distribution of importance values within each category to be visualised. Data points are jittered along the x-axis to reduce overlap. It is worth mentioning that the features shown are those utilised within a multitask predictive approach, in which a single input set is processed through shared layers of the ANN and then used to simultaneously generate forecasts for both electricity price and property demand.}
\label{fig_SHAP_by_method}
\end{figure}


Nevertheless, the magnitude of AFE's influence varied across test set days, yet no instances of AFE-induced performance degradation were observed across any method. For example, adding AFE to the SPO$^{+}$ method improved its performance, leading to regret reductions of 83.22\% and 62.2\% on February 13 and 20, respectively. However, on some days, the improvements were less pronounced. For instance, on 23 February, SPO$^{+}$ improved by 17.1\% (from 0.3367 to 0.2791), indicating that while AFE consistently enhances performance (i.e., yielding lower regrets than without AFE), its effectiveness may vary depending on the relevance of the features generated to the underlying data patterns.

Beyond average improvements, AFE also enhanced the consistency and reliability of performance across all methods. The standard deviations of regrets decreased for all approaches with AFE: from 0.2055 to 0.1732 for PTO (15.72\% reduction), from 0.1206 to 0.0780 for SPO$^{+}$ (35.32\% reduction), and from 0.2474 to 0.2128 for DBB (13.9\% reduction). Notably, SPO$^{+}$ achieved the largest proportional variance reduction from AFE, indicating potential suitability for BESS optimisation problems where data scarcity heightens the importance of methods that respond strongly to AFE.


SHAP values (SHapley Additive exPlanations) were used to interpret the relationships learned by the PTO and DFL methods for predicting electricity price and property demand simultaneously. Across all feature combinations, SHAP computes each input’s average marginal effect on model outputs (i.e., predictions), thereby quantifying both the magnitude and the direction (positive or negative) of feature influence \cite{lundberg2017unified}. As shown in \autoref{fig_SHAP_by_method}, features generated through AFE demonstrate a varying impact on model outputs across different test days. While engineered features dominated the importance rankings on certain days (e.g., 17 and 18 February), raw features such as weather data maintained higher importance scores on others (e.g., 14 and 22 February). Within the AFE categories, statistical nested rolling-window and lag features demonstrate higher impact scores compared to temporal features (e.g., hour of day and its sine/cosine transformations). This may reflect autocorrelation in electricity demand time series, particularly building thermal inertia \cite{MartnezComesaa2020}, where past usage influences near-future demand through gradual thermal changes, making historical patterns more predictive than calendar-based patterns for next-day forecasting. However, feature importance analysis is dataset-dependent, so such observations are likely to vary across different datasets and environments. Yet, a key insight can be drawn that in real-world BESS optimisation scenarios where additional exogenous features are unavailable (e.g., when weather data are not available), prediction performance can still be improved by generating richer inputs from timestamp and historical data alone through AFE. 



\begin{figure}[!t]
\centering
\includegraphics[width=1\linewidth]{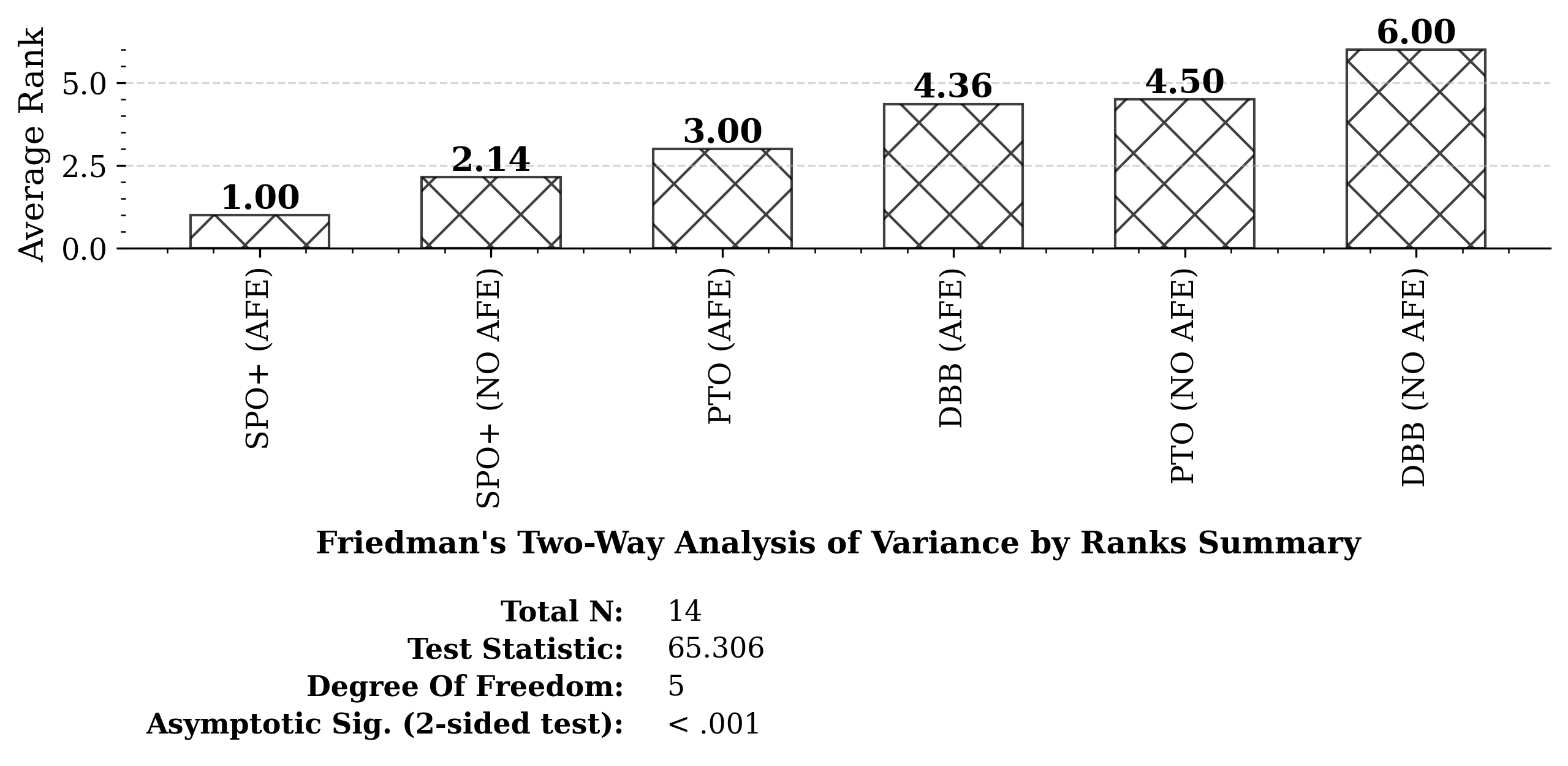}
\caption{Result of comparisons on the test set between the PTO and DFL (SPO$^{+}$ and DBB) methods, with and without AFE, using the Friedman average ranking (lower is better). The significance level is $0.05$.}
\label{Friedman_Ranking_DFL_paper}
\end{figure}



\begin{table}[!t]
\centering
\footnotesize
\setlength{\extrarowheight}{1pt}
\begin{tabular}{@{} p{6cm} c c c c @{}} 
\toprule
Method 1 vs Method 2 & Test Statistic & Std. Test Statistic & Sig. & Adj. Sig.$^{\text{a}}$ \\
\midrule
SPO$^{+}$ (AFE) vs SPO$^{+}$ (No AFE) & -1.143 & -1.616 & .106 & 1.000 \\
SPO$^{+}$ (AFE) vs PTO (AFE) & 2.000 & 2.828 & .005 & .070 \\
SPO$^{+}$ (AFE) vs DBB (AFE) & -3.357 & -4.748 & $<$ .001 & .000 \\
SPO$^{+}$ (AFE) vs PTO (No AFE) & 3.500 & 4.950 & $<$ .001 & .000 \\
SPO$^{+}$ (AFE) vs DBB (No AFE) & -5.000 & -7.071 & $<$ .001 & .000 \\
SPO$^{+}$ (No AFE) vs PTO (AFE) & 0.857 & 1.212 & .225 & 1.000 \\
SPO$^{+}$ (No AFE) vs DBB (AFE) & -2.214 & -3.131 & .002 & .026 \\
SPO$^{+}$ (No AFE) vs PTO (No AFE) & 2.357 & 3.334 & $<$ .001 & .013 \\
SPO$^{+}$ (No AFE) vs DBB (No AFE) & -3.857 & -5.455 & $<$ .001 & .000 \\
PTO (AFE) vs DBB (AFE) & -1.357 & -1.919 & .055 & .824 \\
PTO (AFE) vs PTO (No AFE) & -1.500 & -2.121 & .034 & .508 \\
PTO (AFE) vs DBB (No AFE) & -3.000 & -4.243 & $<$ .001 & .000 \\
DBB (AFE) vs PTO (No AFE) & 0.143 & 0.202 & .840 & 1.000 \\
DBB (AFE) vs DBB (No AFE) & -1.643 & -2.323 & .020 & .302 \\
PTO (No AFE) vs DBB (No AFE) & -1.500 & -2.121 & .034 & .508 \\
\bottomrule
\end{tabular}
\caption{Result of pairwise comparisons on the test set between the PTO and DFL (SPO$^{+}$ and DBB) methods, with and without AFE, using the Wilcoxon signed-rank test. Each row tests the null hypothesis that the Sample 1 (i.e., Method 1) and Sample 2 (i.e., Method 2) distributions are the same. Asymptotic significances (2-sided tests) are displayed. The significance level is $0.05$.\\
\footnotesize{
$^a$ Significance values have been adjusted by the Bonferroni correction for multiple tests.}}
\label{tab:pairwise_comparisons}
\end{table}
\subsection{Statistical Significance Analysis}
\label{Statistical_Significance}
As shown in \autoref{Friedman_Ranking_DFL_paper}, Friedman's test applied to the fourteen-day test horizon reveals a statistically significant global difference among the six methods ($p < .001$), providing compelling evidence that some of the observed performance variations reflect genuine methodological distinctions in handling BESS optimisation under data scarcity rather than random chance. The mean ranks demonstrate that methods with AFE occupy the top three positions out of the top four ranks, with SPO$^{+}$ achieving the best rank, suggesting the superior decision-making efficacy of DFL approaches when combined with AFE for BESS optimisation problems.

Pairwise comparisons using the Wilcoxon signed‑rank test with Bonferroni correction, as shown in \autoref{tab:pairwise_comparisons}, pinpoint where differences are statistically meaningful. SPO$^{+}$ (AFE) demonstrates statistically significant superiority over multiple competing methods. Most notably, SPO$^{+}$ (AFE) significantly outperforms both DBB variants and the PTO approach without AFE (adjusted $p < .001$), reflecting substantial and consistent reductions in regret across the test period. Intriguingly, differences between SPO$^{+}$ (AFE) and PTO (AFE) or between SPO$^{+}$ (AFE) and its own no‑AFE baseline are not significant after correction (adjusted $p=0.070$ and $1.000$, respectively), suggesting that the superiority of SPO$^{+}$ (AFE) over PTO (AFE) or improvements from AFE alone cannot be firmly established (i.e., are not statistically robust under correction). None of the within‑method comparisons (i.e., PTO (AFE) vs PTO (No AFE), DBB (AFE) vs DBB (No AFE), or SPO$^{+}$ (AFE) vs SPO$^{+}$ (No AFE)) remain significant after Bonferroni adjustment, despite unadjusted $p$‑values below 0.05. Overall, the conservative Bonferroni correction reduces the power to detect within-method AFE benefits, though SPO$^{+}$ maintains consistent performance regardless of feature enhancement. This may reflect that SPO$^{+}$ appears less dependent on AFE than alternative approaches, while still benefiting from enhanced feature representations when available. However, it is worth mentioning that the limited test horizon of fourteen days, while sufficient to detect strong global differences, constrains the statistical power to identify more subtle pairwise differences, particularly after stringent multiple comparison corrections.

\section{Conclusion, Limitations and Future Work}
\label{ch:Conclusion}

This work addresses gaps that challenge effective BESS optimisation, stemming from: A) DFL methods promise to align prediction with downstream objectives \cite{Wilder2019, Mandi2020}; however, they are relatively new and have been tested primarily on synthetic datasets or small-scale problems (i.e., simplified benchmarks) \cite{mandi2023towards, Zhou2024}, highlighting the need to assess their practical viability in real-world applications such as BESS problems; and B) real-world datasets often exhibit greater variability and data scarcity due to practical constraints \cite{grinsztajn2022tree, hollmann2022tabpfn} which can compromise DFL performance and necessitate enhanced feature representations to extract richer information from limited data without the need for domain expertise. This work proposes a decision-aware, end-to-end ML framework that directly addresses each gap: first, to overcome data scarcity limitations, the framework leverages domain-specific AFE \cite{Alkhulaifi2025} to extract richer representations without heavy reliance on domain expertise; second, rather than treating forecasting and optimisation as separate tasks, the framework jointly learns to forecast electricity demand and prices while optimising battery operations using a regret-based objective, ensuring prediction errors directly inform decision quality; and third, the study evaluates this framework using a novel dataset collected from a UK household property, providing empirical assessment and evidence of the practical viability of DFL methods in real-world BESS applications.

In this framework, two DFL methods (SPO$^{+}$ and DBB) and a PTO approach are compared under baseline and AFE-enhanced conditions. The results show that AFE improves the performance of all three methods, with particularly pronounced benefits for the DFL approaches (SPO$^{+}$: 56.48\%, DBB: 22.87\%, PTO: 19.42\%). Statistical analysis further confirms significant global differences among methods ($p < .001$), with SPO$^{+}$ (AFE) ranking the best (i.e., lowest regret), demonstrating its superior decision-making efficacy for BESS optimisation under data scarcity. The significance of this work extends beyond the proposed framework and methodological comparisons to practical implications for energy management systems, especially pertinent for real-world BESS deployments, where acquiring large, high-quality datasets is difficult and manual FE is time-consuming and error-prone. The findings indicate that in energy cost-aware scheduling, where future demand and prices are uncertain, aligning prediction and optimisation via DFL’s regret-based training with AFE yields tangible economic benefits through more effective charge/discharge schedules. However, the study also shows that some DFL methods may slightly underperform conventional PTO (i.e., DBB has higher regrets, on average, than PTO), underscoring that sophisticated, emerging learning paradigms alone do not guarantee superior decision quality in data-scarce, real-world BESS applications. It is worth noting that while the problem formulation presented in \autoref{ch:Method}, particularly the mathematical formulation in \autoref{Mathematical_model}, is tailored specifically for BESS applications, the overarching framework methodology is adaptable to other energy management contexts. The core principles of jointly forecasting uncertain parameters while optimising operational decisions remain applicable, though the specific constraints, decision variables, and objective functions would require modification for different energy systems.

Despite the promising results, certain limitations warrant consideration. The study's relatively small dataset (55 days) may limit the generalisability of findings across different energy markets and consumption patterns. Furthermore, hyperparameter optimisation was constrained to a limited search space due to computational resource limitations, as detailed in \autoref{tab:hyperparameters_space}, potentially leaving performance gains unexplored that more extensive tuning might reveal. Future research should explore several promising directions to build upon the current study's findings. First, extending the framework to incorporate renewable generation sources (e.g.,  solar panels) would provide a more comprehensive assessment of DFL's potential in integrated BESSs. Additionally, exploring other DFL approaches, such as noise-contrastive estimation \cite{Mulamba2021}, may yield further improvements. Moreover, testing the framework across longer forecasting horizons, beyond the current one-day ahead predictions, would strengthen the evidence for DFL's applicability in real-world BESS operations. Finally, while the current regret metric focuses solely on cost minimisation, future work could incorporate multi-objective optimisation frameworks that balance economic objectives with environmental considerations, such as carbon emissions reduction, or operational constraints, such as user comfort requirements. 

\section*{CRediT authorship contribution statement}
Nasser Alkhulaifi: Conceptualization, Methodology, Software, Validation, Data Curation, Formal analysis, Investigation, Visualization, Writing – original draft.
Ismail Gokay Dogan: Software, Methodology, Writing – review \& editing.
Timothy R. Cargan:  Methodology, Writing – review \& editing.
Alexander L. Bowler: Conceptualization, Methodology, Writing – review \& editing, Supervision.
Direnc Pekaslan: Writing – review \& editing, Supervision.
Nicholas J. Watson: Conceptualization, Methodology, Writing – review \& editing, Supervision, Funding acquisition.
Isaac Triguero: Conceptualization, Methodology, Validation, Investigation, Writing – review \& editing, Supervision.
\section*{Acknowledgements}
The authors gratefully acknowledge \href{https://www.intelligentplant.com/}{Intelligent Plant Ltd} (Steve Aitken, Josh Plumbly and Bruce Nicolson) for generously providing the battery and demand data used in this study. This work was supported by U.K. Engineering and Physical Sciences Research Council (EPSRC) for the University of Nottingham under Grant EP/S023305/1; in part by the EPSRC Centre for Doctoral Training in Horizon: Creating Our Lives in Data. This work was supported by Grant PID2023-149128NB-I00 funded by MICIU/AEI /10.13039/501100011033 and by ERDF, EU.

\clearpage

\appendix
\gdef\thesection{\Alph{section}}
\makeatletter
\renewcommand\@seccntformat[1]{\appendixname\ \csname the#1\endcsname.\hspace{0.5em}}
\makeatother

\clearpage
\section{Results of the Optimal BESS Scheduling Strategy Across the Test Set Days}
\label{Appendix_A}

\begin{figure}
\centering
\includegraphics[width=1\linewidth]{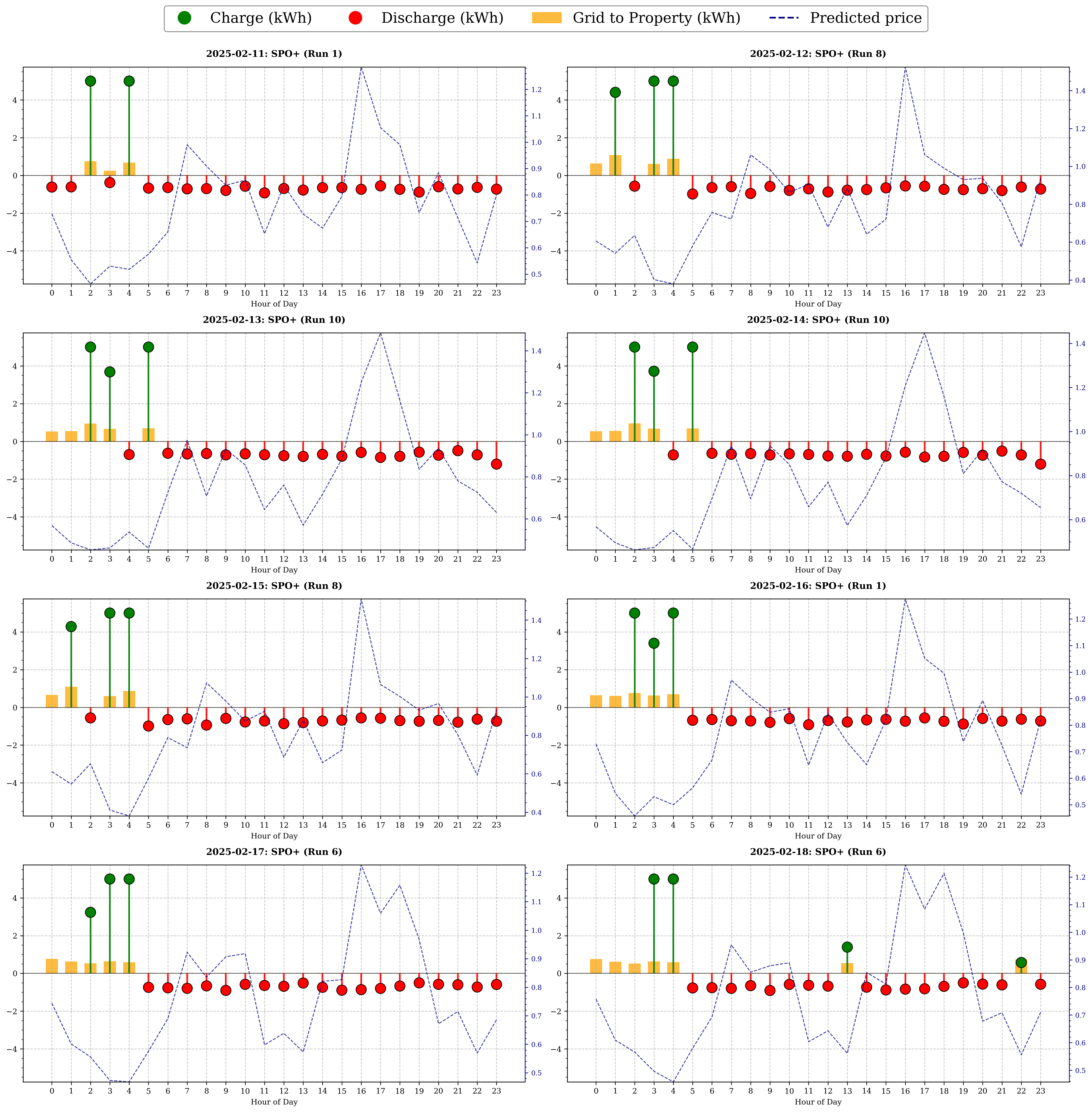}
\caption{Daily BESS optimisation results showing best-performing methods per day. Green circles indicate battery charging, red circles show discharging, and orange background bars represent direct grid supply. The dashed blue line shows predicted electricity prices (right y-axis).}
\label{Scheduling_Strategy_1}
\end{figure}
\begin{figure}[!t]
\centering
\includegraphics[width=1\linewidth]{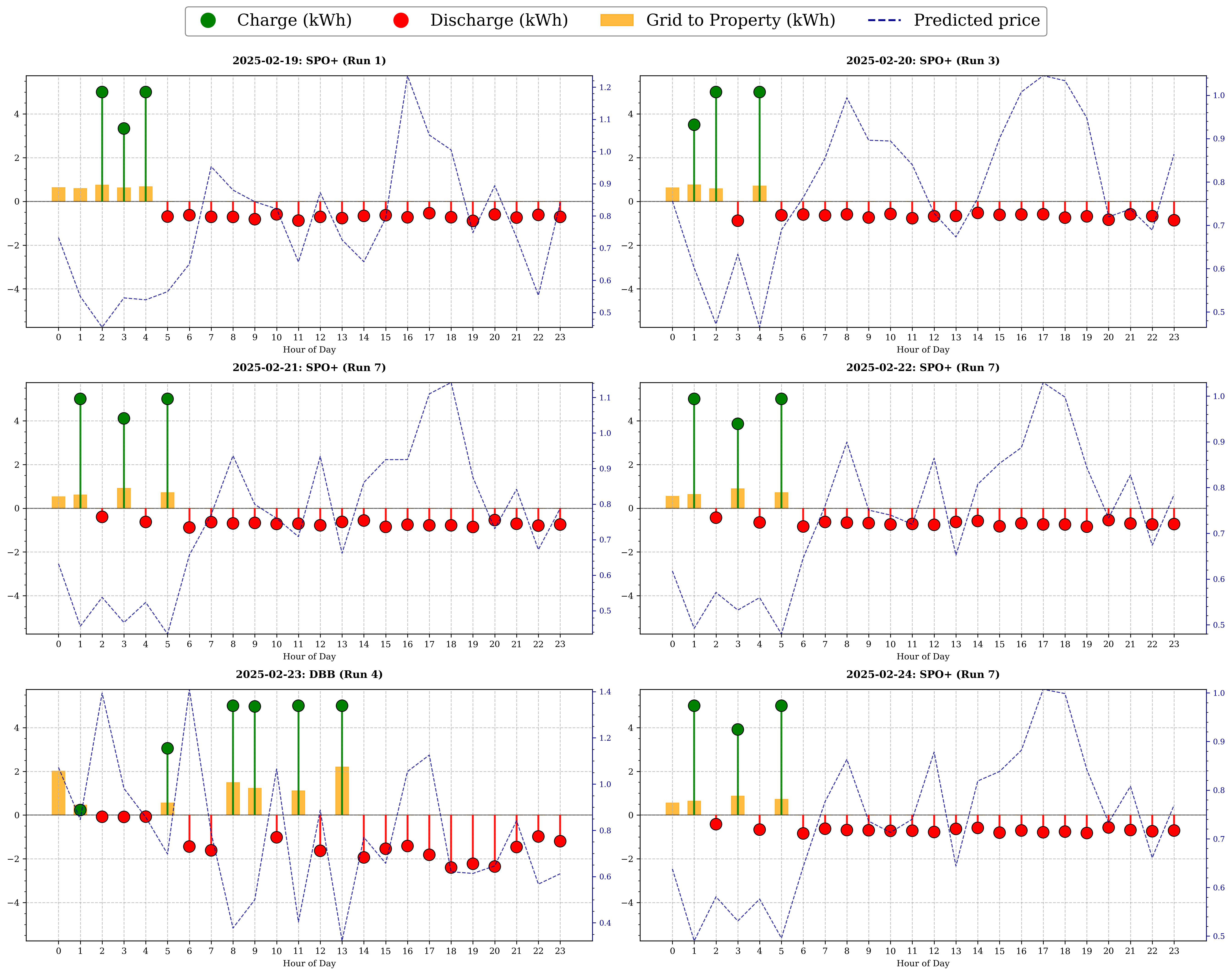}
\caption{Daily BESS optimisation results showing best-performing methods per day. Green circles indicate battery charging, red circles show discharging, and orange background bars represent direct grid supply. The dashed blue line shows predicted electricity prices (right y-axis).}
\label{Scheduling_Strategy_2}
\end{figure}

The following figures present daily BESS optimisation results from multiple experimental runs over all test set days, with each subplot showing the best-performing method for that day based on the lowest regret values. As shown in \autoref{Scheduling_Strategy_1},  and \autoref{Scheduling_Strategy_2}, the visualisation uses a lollipop chart design where battery energy flows are represented by coloured circles connected to stems: green circles above the zero line indicate charging periods (energy stored), while red circles below represent discharging periods (energy released to meet property demand). Orange background bars show direct grid-to-property energy supply where needed. Predicted electricity prices are overlaid as dashed blue lines referenced to the secondary y-axis. Each subplot title identifies the test set date, the best-performing method, and the corresponding experimental run. The horizontal time axis spans 24 hours (00-23), enabling assessment of temporal scheduling patterns in response to price signals.

\clearpage

\bibliographystyle{unsrt}  
\bibliography{references}

\end{document}